\documentclass{article}
\usepackage{cite}
\usepackage{url}
\usepackage[T1]{fontenc}
\usepackage[latin9]{inputenc}
\usepackage{color}
\usepackage{array}
\usepackage{mathtools}
\usepackage{multirow}
\usepackage{amssymb}
\usepackage{setspace}
\usepackage{array}
\date{}
\newcolumntype{L}{>{\centering\arraybackslash}m{3cm}}

\makeatletter

\begin{document}

\title{No reference image quality assessment metric based on regional mutual
information among images}

\newcommand*\samethanks[1][\value{footnote}]{\footnotemark[#1]}
\author{Vinay Kumar\thanks{Department of Electronics \& Communication Engineering, Thapar Institute of Engineering and Technology,
Patiala}
\and Vivek Singh Bawa\thanks{Visual Artificial Intelligence Laboratory, Oxford Brookes University, UK.}
\and Rahul Upadhyay\samethanks[1]}

\maketitle
\begin{abstract}
With the inclusion of camera in daily life, an automatic no reference
image quality evaluation index is required for automatic classification
of images. The present manuscripts proposes a new No Reference Regional
Mutual Information based technique for evaluating the quality of an
image. We use regional mutual information on subsets of the complete
image. Proposed technique is tested on four benchmark natural image
databases, and one benchmark synthetic database. A comparative analysis
with classical and state-of-art methods indicate superiority of the
present technique for high quality images and comparable for other
images of the respective databases. 
\end{abstract}

\section{Introduction}

With the advent of inexpensive and good quality mobile cameras storage,
transmission and compression of images has become a standard practice
among technical and non technical masses. Large number of people have
mobile phones with camera capturing trillions of photographs every
year \cite{art1}, approximately 24 billion selfies were uploaded
to Google in year 2015 \cite{art2} and increasing exponentially with
every passing year. Unlimited space for uploading images on Google
photos (and large space on other web servers; for example, Flickr,
Pinterest, etc) facilitates and influences people to capture many
photographs of the same situation. Searching the good quality images
from this ever (exponentially) increasing large quantity is impossible
task for a human being. Therefore, it becomes pertinent to design
and develop better automatic and no-reference image quality assessment
system. These systems will help; for example, in evaluating the image
information and (possibly) retain the best out of plethora, find out
the quality in real time, selecting camera settings for best results,
etc. This drives researchers to develop better auto no reference image
quality measurement techniques \cite{Lin2011,Liu2012,Narwaria2010,Narwaria2012,Liang2010,Wang2013,Wang2002a}. 

Researchers generally talk about three types of image quality assessment
(IQA) techniques: 
\begin{enumerate}
\item full reference \cite{Zhang2011,Li2016,Wu2013,Gu2014,Wang2012,Wang2011a,Zhang2014,Fang2014}, 
\item reduced reference IQA \cite{Tao2009,Gao2009}, and 
\item no-reference IQA \cite{Mittal2012,Gu2015,Gu2016,Zhang2015}. 
\end{enumerate}
First type of IQA assumes that human beings are sensitive to degradations,
second indicates that we are more sensitive to few key features extracted
from the image. The current proposed technique lies in the last category. 

Various techniques for objective image quality measurement are discussed
in literature. Since human visual system is a complex set of decision
making processes, available IQA methods are still not as good as the
human visual decisions. We discuss some of the relevant and prevailing
methods in rest of the present section.

Wu et al \cite{Wu1997} used measurement of blocking effect in horizontal
and vertical directions and differences at block boundaries in horizontal
and vertical directions, respectively. Tan et al \cite{Tan2000a}
analyzes magnitude and phase information in a harmonics to measure
the quality of the image. Another \cite{Tan2000} model was developed
for measuring block effects in an image. Wang et al \cite{Wang2000}
used energy based measurements to find the blocking artifacts in an
image. These blocking effects become fundamental building blocks for
measurement of quality of an image.

While transmitting or storing, image quality (IQ) measurement plays
a crucial role to evaluate and choose the correct image. The ultimate
goal of IQ measurement is assigning a quantitative value to perception
to human observers. Researchers perform this task with the help of
crowd sourcing and acquiring Mean Opinion Score (MOS). MOS or its
modified versions compared with the IQA values become basis for quality
of the IQA index. 

In the next section we discuss proposed method followed, in section
\ref{sec:Experimental-Results}, by experimental results and discussion.
We close the manuscript with conclusion and references. 

\section{\label{sec:Methodology}Methodology}

The proposed index No Reference Regional Mutual Information (NrMI)
predicts the quality of an image with the help of following procedure. 

Given an image matrix $\Phi(x,y)\in\mathbb{Z}^{n\times m}$. Another
version of matrix $\Phi(x,y)$ is created and is depicted by $\Phi'_{\theta}(x',y')$,
where

\begin{equation}
\left[\begin{array}{c}
x'\\
y'
\end{array}\right]=\left[\begin{array}{cc}
cos\theta & -sin\theta\\
sin\theta & cos\theta
\end{array}\right]\left[\begin{array}{c}
x\\
y
\end{array}\right]\label{eq:thetaRot}
\end{equation}

To make $\Phi$ and $\Phi'_{\theta}$ of same size equations \ref{eq:vect}
and \ref{eq:invvect} are applied. 
\begin{equation}
\Phi_{vec}(v)=vec(\Phi'_{\theta}(x',y'))\label{eq:vect}
\end{equation}

\begin{equation}
\Phi_{\theta v}(x,y)=vec^{-1}(\Phi_{vec}(v))\vcentcolon\Phi_{\theta}(x,y)\in\mathbb{R}^{ab}\rightarrow\mathbb{R}^{n\times m}\label{eq:invvect}
\end{equation}

We divide $\Phi(x,y)$ into disjoint group of $n$ sub-matrices, $\eta_{k}^{a\times b}\colon k\in\mathbb{Z^{>}}$where
$\eta_{k}:\eta_{k}\subset\Phi$. $\eta_{k}$ contains $q$ member
of perfect subsets of $\Phi$ (such that $k/m\in\mathbb{Z^{>}}$),
for it is obvious that if a subset is perfect, then there is no information
loss. Every element of $\eta_{k}$ represents a segment of original
image $\Phi$. $\Phi'_{\theta}(x',y')$ is divided into sub-matrices
$\eta_{k,\theta}^{a\times b}$ ($\equiv\eta_{k}$). 

We choose size of $\eta_{k}$ (and consequently $\eta_{k,\theta}$)
to be $3\times3$, which makes sure that the values within $\eta_{k}$
will not be varying significantly except when sub-matrix lies at an
edge in $\Phi(x,y)$ (or $\Phi'_{\theta}(x',y')$). The value of $\theta$
is $\pi/2$, one can choose any value for $\theta$ but $\pi/2$ provides
maximum shift, and equation \ref{eq:thetaRot} is rewritten as

\begin{equation}
\left[\begin{array}{c}
x'\\
y'
\end{array}\right]=\left[\begin{array}{c}
-y\\
x
\end{array}\right]
\end{equation}

which relates image matrices $\Phi$ and $\Phi'_{\theta}$ by equation
\ref{eq:phiRel} 
\begin{equation}
\Phi(x,y)=\Phi'_{\theta}(-y,x)\label{eq:phiRel}
\end{equation}

Let sub-matrices $\eta_{k}$ (and $\eta_{k,\theta}$) be represented
by 

\begin{equation}
\begin{array}{ccc}
\eta_{k} & = & \left[\begin{array}{ccc}
e_{11} & e_{12} & e_{13}\\
e_{21} & e_{22} & e_{23}\\
e_{31} & e_{32} & e_{33}
\end{array}\right]\\
\eta_{k,\theta} & = & \left[\begin{array}{ccc}
e_{11,\theta} & e_{12,\theta} & e_{13,\theta}\\
e_{21,\theta} & e_{22,\theta} & e_{23,\theta}\\
e_{31,\theta} & e_{32,\theta} & e_{33,\theta}
\end{array}\right]
\end{array}\label{eq:elements}
\end{equation}

From matrices of equation \ref{eq:elements} calculate matrix $M_{e}$ 

\begin{equation}
M_{e}=\left[e_{11}\;e_{12}\;e_{13}\;e_{11,\theta}\;e_{12,\theta}\;e_{13,\theta}\;e_{21}\ldots e_{23}\;e_{21,\theta}\;\ldots\;e_{33,\theta}\right]\label{eq:MatrixP}
\end{equation}

Center the values at the origin and represent it by $M_{e,0}$ by
\begin{equation}
M_{e,0}=M_{e}-\frac{1}{N}\sum_{i}^{N}p_{i}
\end{equation}

where $p_{i}$ are elements of matrix $M_{e}$ and $N=9+9=18$.

Find covariance 
\begin{equation}
C=\frac{1}{N}M_{e,0}M_{e,0}^{T}
\end{equation}

Estimate joint entropy \footnote{\label{fn:Joint-and-marginal}Joint and marginal entropy is given
by \cite{Reza1961}
\begin{equation}
H_{g}(\Sigma_{d})=log((2\pi e)^{\frac{d}{2}}det(\Sigma_{d})^{\frac{1}{2}})
\end{equation}
which represents the entropy of a normally distributed set of points
in $\Re^{d}$ with covariance matrix $\Sigma_{d}$.}
\[
H_{g}(C)
\]

Estimate marginal entropy\ref{fn:Joint-and-marginal} $H_{g}(C_{A})$
and $H_{g}(C_{B})$, where $C_{A}$ and $C_{B}$ are top left and
bottom right $\frac{d}{2}\times\frac{d}{2}$ matrices of $C$.  $d$
is a relationship defined as

\begin{equation}
d=2(2r+1)^{2}\label{eq:val-r}
\end{equation}

where, $r$ is the size of sub matrix under consideration; that is,
the size of MB for which we are going to calculate the similarity
within the matrix; for example  in Figure value of $r$ is $1$. 

Calculate Regional Mutual Information
\begin{equation}
M_{rmi}=H_{g}(C_{A})+H_{g}(C_{B})-H_{g}(C)\label{eq:RMI}
\end{equation}

$M_{rmi}$ gives a measure of regional mutual information between
$\Phi(x,y)$ and $\Phi_{\theta}(x',y')$. 

A weight function for RMI is calculated with equation of $\Phi_{vec}(v)$
is calculated next 
\begin{equation}
\Phi_{wg}=\mathbf{E}\left[(\Phi_{vec}(v)-\mathbf{E}[\Phi_{vec}])^{2}\right]\label{eq:varweight}
\end{equation}

The relative quality of an image is given by
\begin{equation}
NrMI_{i}=M_{rmi,i}*\Phi_{wg,i}\label{eq:qualityMat}
\end{equation}

where $i\in i^{th}$ image in the image sequence.

\section{\label{sec:Experimental-Results}Experimental Results }

In this section we validate our method through application on various
benchmark state-of-art and classical databases. Experiments are conducted
with five standard databases of natural and one of synthetic images.
The natural image databases are TID 2008 \cite{Ponomarenko2009} with
1699 images, TID 2013 \cite{Ponomarenko2013} with 2483 images, CID
2013 \cite{Larson2010}, LIVE \cite{Sheikh2006}, MEFD with 550 images
each. While ESPL \cite{Kundu2015}, a database consisting of 550 synthetic
images, is used for evaluation of the current algorithm.

For objective evaluation SRCC (Spearman's Rank Correlation Coefficient)
and PLCC (Pearson Linear Correlation Coefficient) matrices are used.
These metrics give a measure of prediction monotonicity and linearity,
respectively. 

Table \ref{tab:Performance-Comparison-of} presents a comparative
view of various index of quantitative quality measures. Blue color
values in table \ref{tab:Performance-Comparison-of} indicate best
results. Since no-reference quality measurement system requires complex
set of interdependent parameters to work as efficiently as human beings,
therefore every system has certain advantage over others under certain
conditions. From the table it becomes clear that proposed method evaluates
the images better than SSIM for most of the databases. Since the proposed
method uses underlying regional geometric information by splitting
the set into disjoint group of sub-sets; therefore every small change
in geometry (including presence of undetectable noise for human visual
system) changes the qualitative measure. 

Specifically with images of high quality (databases MEFD and ESPL)
the proposed method performs much better than SSIM \cite{Wang2004}
and other state-of-art techniques. Since proposed technique considers
underlying geometry of the image, high quality images distorted by
small amount of noise produce lower value of the quality index. This
lower value in turn will be helpful to take corrective measures to
develop noise removal or better compression algorithms. 

\begin{table*}
\centering

\begin{tabular}{|>{\centering}p{2cm}|>{\centering}p{2cm}|>{\centering}p{1cm}|>{\centering}p{1cm}|>{\centering}p{1cm}|>{\centering}p{1cm}|>{\centering}p{1cm}|>{\centering}p{1cm}|>{\centering}p{1cm}|}
\hline 
{\scriptsize{}Database} & {\scriptsize{}Statistical Measurement} & {\scriptsize{}SSIM }\cite{Wang2004}{\scriptsize{} } & \multirow{1}{1cm}{{\scriptsize{}NR \cite{Wang2002}}} & {\scriptsize{}NJQA \cite{Golestaneh2014}} & {\scriptsize{}NR \cite{Li2015}} & {\scriptsize{}MUG NR \cite{Nafchi2016}} & {\scriptsize{}MUG$^{+}$ NR \cite{Nafchi2016}} & {\scriptsize{}PM}\tabularnewline
\hline 
\hline 
{\scriptsize{}TID 2008 \cite{Ponomarenko2009}} & {\scriptsize{}PLCC} & \textcolor{blue}{\scriptsize{}0.954} & {\scriptsize{}0.952} & {\scriptsize{}0.944} & \textcolor{black}{\scriptsize{}0.951} & {\scriptsize{}0.941} & \textcolor{black}{\scriptsize{}0.953} & {\scriptsize{}0.868}\tabularnewline
\hline 
 & {\scriptsize{}SRCC} & \textcolor{blue}{\scriptsize{}0.925} & {\scriptsize{}0.913} & {\scriptsize{}0.8993} & {\scriptsize{}0.917} & \textcolor{black}{\scriptsize{}0.917} & \textcolor{black}{\scriptsize{}0.924} & {\scriptsize{}0.832}\tabularnewline
\hline 
{\scriptsize{}TID 2013 \cite{Ponomarenko2013}} & {\scriptsize{}PLCC} & \textcolor{black}{\scriptsize{}0.954} & {\scriptsize{}0.953} & {\scriptsize{}0.948} & \textcolor{black}{\scriptsize{}0.955} & {\scriptsize{}0.942} & \textcolor{blue}{\scriptsize{}0.955} & {\scriptsize{}0.887}\tabularnewline
\hline 
 & {\scriptsize{}SRCC} & {\scriptsize{}0.9200} & \textcolor{black}{\scriptsize{}0.927} & {\scriptsize{}0.886} & \textcolor{blue}{\scriptsize{}0.931} & {\scriptsize{}0.908} & {\scriptsize{}0.919} & {\scriptsize{}0.842}\tabularnewline
\hline 
{\scriptsize{}CID 2013 \cite{Larson2010}} & {\scriptsize{}PLCC} & \textcolor{black}{\scriptsize{}0.979} & \textcolor{black}{\scriptsize{}0.975} & {\scriptsize{}0.954} & \textcolor{blue}{\scriptsize{}0.979} & {\scriptsize{}0.9679} & {\scriptsize{}0.972} & {\scriptsize{}0.789}\tabularnewline
\hline 
 & {\scriptsize{}SRCC} & {\scriptsize{}0.955} & \textcolor{black}{\scriptsize{}0.955} & {\scriptsize{}0.925} & \textcolor{blue}{\scriptsize{}0.957} & {\scriptsize{}0.930} & {\scriptsize{}0.937} & {\scriptsize{}0.798}\tabularnewline
\hline 
{\scriptsize{}LIVE \cite{Sheikh2006}} & {\scriptsize{}PLCC} & \textcolor{blue}{\scriptsize{}0.979} & \textcolor{black}{\scriptsize{}0.979} & {\scriptsize{}0.956} & {\scriptsize{}0.976} & {\scriptsize{}0.965} & {\scriptsize{}0.973} & {\scriptsize{}0.962}\tabularnewline
\hline 
 & {\scriptsize{}SRCC} & {\scriptsize{}0.946} & \textcolor{blue}{\scriptsize{}0.974} & {\scriptsize{}0.956} & \textcolor{black}{\scriptsize{}0.973} & {\scriptsize{}0.959} & {\scriptsize{}0.968} & {\scriptsize{}0.959}\tabularnewline
\hline 
{\scriptsize{}ESPL \cite{Kundu2015}} & {\scriptsize{}PLCC} & {\scriptsize{}0.943} & {\scriptsize{}0.960} & {\scriptsize{}0.809} & \textcolor{blue}{\scriptsize{}0.962} & {\scriptsize{}0.940} & {\scriptsize{}0.937} & \textcolor{blue}{\scriptsize{}0.962}{\scriptsize{} }\tabularnewline
\hline 
 & {\scriptsize{}SRCC} & {\scriptsize{}0.904} & {\scriptsize{}0.933} & {\scriptsize{}0.739} & \textcolor{black}{\scriptsize{}0.933} & {\scriptsize{}0.928} & {\scriptsize{}0.927} & \textcolor{blue}{\scriptsize{}0.959}\tabularnewline
\hline 
\end{tabular}

\caption{\label{tab:Performance-Comparison-of}Performance comparison of no
reference image quality measure for TID 2008 \cite{Ponomarenko2009},
TID 2013 \cite{Ponomarenko2013}, CID 2013 \cite{Larson2010}, LIVE
\cite{Sheikh2006}, MEFD, and ESPL \cite{Kundu2015} databases. Blue
color values represent best performing technique in terms of corresponding
SRCC and PLCC statistical measures. }
\end{table*}

\section{\label{sec:Conclusion}Conclusion}

Present manuscript investigates the problem of no-reference quality
assessment. A novel technique has been proposed for the assessment
based on underlying geometry of the image. The technique is applied
on various databases with different types of images. Results show
interesting trend and promising performance when compared with existing
literature. Since method utilizes mutual information approach it was
able to render better results for high quality images. 

In future we aim to study the effects of current technique by calculating
RMI on weighted image segments. The weights will be calculated based
on the importance of the region in the images, which in turn depends
on point of focus in human visual system. 

\bibliographystyle{plain}
\bibliography{myreferences}

\end{document}